\pdfoutput=1

\documentclass[11pt]{article}

\usepackage{EMNLP2023}

\usepackage{times}
\usepackage{latexsym}
\usepackage{adjustbox}
\usepackage[T1]{fontenc}
\usepackage{float}
\usepackage{multicol}
\usepackage{hyperref}
\usepackage{times}
\usepackage{algorithm}
\usepackage{latexsym}
\usepackage{url}
\usepackage{array}
\usepackage{booktabs}
\usepackage{stfloats}
\usepackage{algpseudocode}
\usepackage{float}
\usepackage{multirow}
\usepackage{adjustbox}
\usepackage{makecell}
\usepackage{graphicx}
\usepackage{caption}
\usepackage{subcaption}
\usepackage{xcolor}
\usepackage{appendix}
\usepackage{enumitem}
\usepackage{threeparttable}
\usepackage{tcolorbox} 
\tcbuselibrary{skins,breakable}
\usepackage{listings}
\usepackage{tabularx}
\usepackage{amssymb}
\usepackage[table]{xcolor}
\usepackage[utf8]{inputenc}

\usepackage{microtype}

\usepackage{inconsolata}

\title{Judge Model for Large-scale Multimodality Benchmarks}

\author{
Min-Han Shih,
Yu-Hsin Wu,
Yu-Wei Chen \\
Department of Electrical and Computer Engineering, \\ 
Viterbi School of Engineering, University of Southern California, United States \\
\texttt{\{minhansh, yuhsinwu, ychen543\}@usc.edu}
}
\begin{document}
\maketitle

\begin{abstract}
We propose a dedicated multimodal Judge Model designed to provide reliable, explainable evaluation across a diverse suite of tasks. Our benchmark spans text, audio, image, and video modalities, drawing from carefully sampled public datasets with fixed seeds to ensure reproducibility and minimize train–test leakage. Instead of simple scoring, our framework aggregates multimodal judgments, analyzes the quality and reasoning consistency of model outputs, and generates diagnostic feedback. We evaluate state-of-the-art MLLMs, including Gemini-2.5-flash, Phi-4, and Qwen-2.5-omni, across 280 multimodal samples and compare judge-model assessments with human annotators. Results show strong alignment between the Judge Model and human scores, demonstrating its potential as a scalable, interpretable evaluation pipeline for future multimodal AI research. 
\end{abstract}

\section{Introduction}
Recent advances in Large Language Models (LLMs) and Multimodal Large Language Models (MLLMs) have demonstrated remarkable progress in understanding and reasoning across multiple modalities such as text, image, audio, and video. Models like GPT-5 , Gemini 2.5, and Claude 4 sonnet can now interpret complex multimodal inputs, perform reasoning over long contexts, and generate structured, contextually aware outputs. However, as the capability and diversity of MLLMs expand, objectively and efficiently evaluating their performance has become an increasingly challenging problem. 

Traditional evaluation pipelines rely on either human annotation or static benchmark scoring, which are often time-consuming, inconsistent, and limited to task-specific performance. Meanwhile, recent approaches, such as LLM-as-a-Judge, leverage large models as evaluators. \cite{llmjudge, llmjudge2, llmjudgeQA} Although these methods reduce human workload, they introduce several issues: (1) lack of multimodal understanding, since many judge systems are purely text-based; and (2) insufficient explainability and reliability when compared to human judgment on complex reasoning or cross-modal alignment tasks. 

To address these limitations, our project proposes a \textbf{Judge Model} designed specifically for large-scale multimodal benchmarks. Instead of scoring, our framework aggregates and calibrates judgments across multiple perspectives. The Judge will evaluate the outputs of various MLLMs, such as Gemini 2.5, Phi-4 \cite{phi}, and Qwen-2.5 \cite{qwen2025qwen25technicalreport}, analyzing the quality of the response, the alignment of the modality and the consistency of the reasoning trace. Beyond scoring, it will also generate comprehensive feedback and guidance, aiming to establish an interpretable and reliable evaluation pipeline for multimodal AI systems.

\section{Related Work}
\subsection{Reasoning LLMs}
LLMs know when to think step-by-step and only generate step-by-step reasoning when necessary. \cite{wei2023chainofthoughtpromptingelicitsreasoning}  Prior works on Chain-of-Thought (CoT) mostly focus on the correctness of the CoT reasoning steps or whether the reasoning steps are faithful to the question and support the final answer. \cite{ye2022unreliabilityexplanationsfewshotprompting, golovneva2023roscoesuitemetricsscoring} 
\subsection{LLM-as-the-Judge}
LLM-based evaluators are LLMs that are prompted to judge the quality of some samples based on specific criteria. LLM-based evaluators evaluate the quality of a single sample using a score such as Likert scores \cite{likert} on a scale of 1 to 5. \cite{chiang-etal-2024-large, chiang-lee-2023-closer, chiang-lee-2023-large, chiang-lee-2024-reasoning, zheng2023judgingllmasajudgemtbenchchatbot} LLM-based evaluators can also compare the quality of a pair of samples and judge which one is better. \cite{zheng2023judgingllmasajudgemtbenchchatbot}

\subsection{Multimodal Benchmark}
Vision benchmark expands from image benchmarks to video benchmarks that introduce temporal structure and richer scene context \cite{nguyen2025seehearunderstandbenchmarking}, such as
Vision Quesion-Answering \cite{vqa, vqa2}, chart understanding \cite{chartqa, chartinggapsrealistic, documentqa}, and general capability suites \cite{fu2025mmecomprehensiveevaluationbenchmark, li2023seedbenchbenchmarkingmultimodalllms, liu2024mmbenchmultimodalmodelallaround}. Different video datasets emphasize different abilities: temporal ordering \cite{temporal, videollmsreally}, procedural understanding \cite{coin, xiao2021nextqanextphasequestionansweringexplaining}, egocentric perception \cite{ego4d, damen2018scalingegocentricvisionepickitchens}, and world modeling \cite{hong2025worldsenseevaluatingrealworldomnimodal}.  Audio benchmarks such as AudioSet \cite{audiocaps} and VGGSound \cite{vggsound} target event-level acoustic classification, while speech datasets cover specific facets of communication: VoxCeleb \cite{vox, vox2} for speaker identity, AudioCaps / SpeechCaps \cite{audiocaps, SpeechCaps} for audio/speech captioning, LRS \cite{afouras2018lrs3tedlargescaledatasetvisual} for transcription, and AVA-ActiveSpeaker \cite{kim21k_interspeech} for frame-level speaking and audibility labels. Dynamic-SUPERB \cite{huang2024dynamicsuperbdynamiccollaborativecomprehensive, huang2025dynamicsuperbphase2collaborativelyexpanding} proposed a dynamic, collaborative benchmark in speech and audio tasks.

\section{Datasets}
\subsection{Text-to-Text Datasets}
To build a compact, reproducible benchmark, we pull small, fixed-size samples from widely used public datasets on HuggingFace Hub. For each dataset we explicitly name the split used so that readers can assess the risk of train--test leakage in future model training. We sample without replacement using a fixed random seed. The data source and the number of each task are listed in Table \ref{tab:tt_datasets}. 


\begin{table}[h]
\centering
\resizebox{\linewidth}{!}{
    \begin{tabular}{l l l r l}
    \toprule
    \textbf{Task Family} & \textbf{Dataset (HF ID)} & \textbf{Split} & \textbf{\#Items} & \textbf{Answer Type}\\
    \midrule
    Reasoning--Code & \texttt{nvidia/OpenCodeReasoning} & \texttt{split\_0} & 20 & code / final string \\
    Reasoning--Math & \texttt{nvidia/OpenMathReasoning} & \texttt{cot} & 20 & generative (final value) \\
    Expert MCQ & \texttt{Idavidrein/gpqa} (\emph{fallback:} \texttt{fingertap/GPQA-Diamond}) & \texttt{test} & 10 & multiple choice \\
    Reading Comprehension & \texttt{ucinlp/drop} & \texttt{validation} & 20 & extractive span \\
    Commonsense Reasoning & \texttt{Rowan/hellaswag} & \texttt{validation} & 15 & multiple choice \\
    Commonsense Reasoning & \texttt{jet-ai/social\_i\_qa} & \texttt{validation} & 15 & multiple choice \\
    Instruction Following & \texttt{google/IFEval} & \texttt{train} & 15 & generative (constrainted) \\
    Instruction Following & \texttt{YuxinJiang/FollowBench} & \texttt{train} (\emph{fallback:} \texttt{validation}) & 15 & generative (constrainted) \\
    \midrule
    \multicolumn{2}{l}{\textbf{Total}} & & \textbf{130} & \\
    \bottomrule
    \end{tabular}
}
\caption{Datasets, splits, counts, and answer types used in our text benchmark.}
\label{tab:tt_datasets}
\end{table}
\vspace{-1.5em}

\subsection{Text-to-Text Tasks} \label{sec:tt_tasks}
1 .\textbf{Reasoning--Code:} program synthesis problem solving requiring structured outputs or final values. \cite{ocr} \\
2.  \textbf{Reasoning--Math:} multi-step mathematical reasoning aiming at a final canonical answer.  \cite{omr} \\
3. \textbf{Expert MCQ:} graduate-level STEM multiple-choice questions \cite{gpqa, crossfit, sys}. \\
4.  \textbf{Reading Comprehension:} passage-based QA with extractive answers \cite{DROP} \\
5. \textbf{Commonsense Reasoning:} everyday physical/social reasoning in multiple-choice format. \cite{hellaswag, socialIQa} \\
6. \textbf{Instruction Following:} adherence to explicit constraints and formatting instructions. \cite{followbench, ifeval}

\subsection{Audio-Language Datasets} \label{sec:al_tasks}
Our audio-language datasets are constructed from four open-source datasets hosted on Hugging Face. These datasets were carefully selected to ensure diversity across audio modalities and task types, 
including environmental sound understanding, bird sound classification, music generation, and speech question answering. The sources and task types are summarized in Table \ref{tab:audio}.

\begin{table}[h]
\centering
\resizebox{\linewidth}{!}{
\begin{tabular}{l l l r}
    \toprule
    \textbf{Task Family} & \textbf{Dataset (HF ID)} & \textbf{Split} & \textbf{\#Items} \\
    \midrule
    Audio Captioning & \texttt{kuanhuggingface/audiocaps\_hallucination} & \texttt{test} & 10 \\
    Birdsound Detection & \texttt{tglcourse/5s\_birdcall\_samples\_top20} & \texttt{validation} & 10 \\
    Music Style Detection & \texttt{nyuuzyou/suno} & \texttt{train} & 10 \\
    Speech Question-Answering & \texttt{AudioLLMs/public\_sg\_speech\_qa\_test} & \texttt{test} & 20 \\
    \midrule
    \textbf{Total} & & & \textbf{50} \\
    \bottomrule
\end{tabular}
}
\caption{Datasets, splits, counts, and answer types used in our audio benchmark.}
\label{tab:audio}
\end{table}
\vspace{-1.5em}

\subsection{Audio-Language Tasks} \label{sec:al_tasks}
1. \textbf{Question–Answer:} Involve recognizing and describing the presence or absence of specific sounds, or answering questions based on audio content. \cite{gsqa, audiocaps, audiobench}\\
2. \textbf{Instruction–Following:} Evaluate the model’s ability to interpret and execute textual instructions for auditory inputs. \cite{audiobench}

\subsection{Video Datasets} \label{sec:vl_tasks}
Our video dataset comes from the Hugging Face public dataset. It consists of AI-generated short clips created by generative video models such as Sora, Veo2, and Kling. Each clip is paired with human-curated questions designed to test the reasoning of a model about common sense and physical consistency. To ensure reproducibility, we sample using a fixed random seed and select a total of 50 video clips from non-training sets, which minimizes the chance of overlap with materials that LLMs might have already seen during pretraining. This subset provides a controlled and diverse evaluation set for testing video content understanding and hallucination robustness in MLLMs. The sources and task types are summarized in Table \ref{tab:video}.
\begin{table}[H]
\centering
\resizebox{\linewidth}{!}{
    \begin{tabular}{@{} l l r @{}}
    \toprule
    \textbf{Dataset (HF ID)} & \textbf{Split} & \textbf{\#Items}\\
    \midrule
    \texttt{IntelligenceLab/VideoHallu} & \texttt{validation} & 25 \\
    \texttt{IntelligenceLab/VideoHallu} & \texttt{test} & 25 \\
    \midrule
    \multicolumn{2}{l}{\textbf{Total}} & \textbf{50} \\
    \bottomrule
    \end{tabular}
}
\caption{Datasets, splits, counts, and answer types used in our video benchmark.}
\label{tab:video}
\end{table}
\vspace{-1.5em}

\subsection{Video Tasks} \label{sec:vl_tasks}
Our task focuses on common sense and physical reasoning rather than traditional visual recognition, due to the use of AI-generated synthetic videos.
The model is required to analyze each short clip, interpret the given question, and determine whether the depicted scene is plausible or abnormal.
By framing the problem in this way, we shift the emphasis from simple visual perception to critical reasoning, allowing us to evaluate how effectively MLLMs can align visual information with their internal understanding of real-world knowledge.

\subsection{Iamge Datasets}\label{sec:vl_tasks}
To construct a diverse and interpretable benchmark for image–language understanding, we curated samples from four public datasets on Hugging Face, each representing a distinct subdomain of visual reasoning. For all datasets, we sample small, fixed-size subsets using a consistent random seed for reproducibility. Only evaluation or test splits are used to minimize potential data leakage from pretraining corpora. The sources and task types are summarized in Table \ref{tab:Image-Dataset-FP}.
\begin{table}[H]
\centering
\resizebox{\linewidth}{!}{
    \begin{tabular}{@{} l l l r @{}}
    \toprule
    \textbf{Task} & \textbf{Dataset (HF ID)} & \textbf{Split} & \textbf{\#Items}\\
    \midrule
    Image Captioning & \texttt{nyu-visionx/CV-Bench} & \texttt{test} & 10 \\
    Image Captioning & \texttt{nirajandhakal/realworldqa} & \texttt{test} & 10 \\
    Chart Understanding& \texttt{ChartFoundation/ECDBench} & \texttt{test} & 15 \\
    Math Reasoning & \texttt{MathLLMs/MathVision} & \texttt{test} & 15 \\
    \midrule
    \multicolumn{3}{l}{\textbf{Total}} & \textbf{50} \\
    \bottomrule
    \end{tabular}
}
\caption{Datasets, splits, counts, and answer types used in our image benchmark.}
\label{tab:Image-Dataset-FP}
\end{table}
\vspace{-1.5em}

\subsection{Iamge Tasks} \label{sec:vl_tasks}
1. \textbf{Image Captioning:} Evaluates the model’s ability to perform commonsense reasoning and holistic scene understanding. It assesses whether the model can generate captions that reflect physical plausibility, spatial and contextual consistency, and realistic visual interpretation. The images span both everyday real-world scenes and complex scenarios that require object recognition, contextual inference, and higher-level reasoning. \cite{cvbench} \\
2. \textbf{Chart Understanding:} Measures quantitative and relational reasoning from structured visuals like bar charts, requiring comparison of trends and numerical relationships. \cite{ecdbench} \\
3. \textbf{Math Reasoning:} Assesses symbolic and spatial reasoning using mathematical diagrams and geometric figures that demand parsing and quantitative inference. \cite{mathv1, mathv2}

\section{Methodology}
\begin{figure*}[t!]
    \centering
    \includegraphics[width=\linewidth]{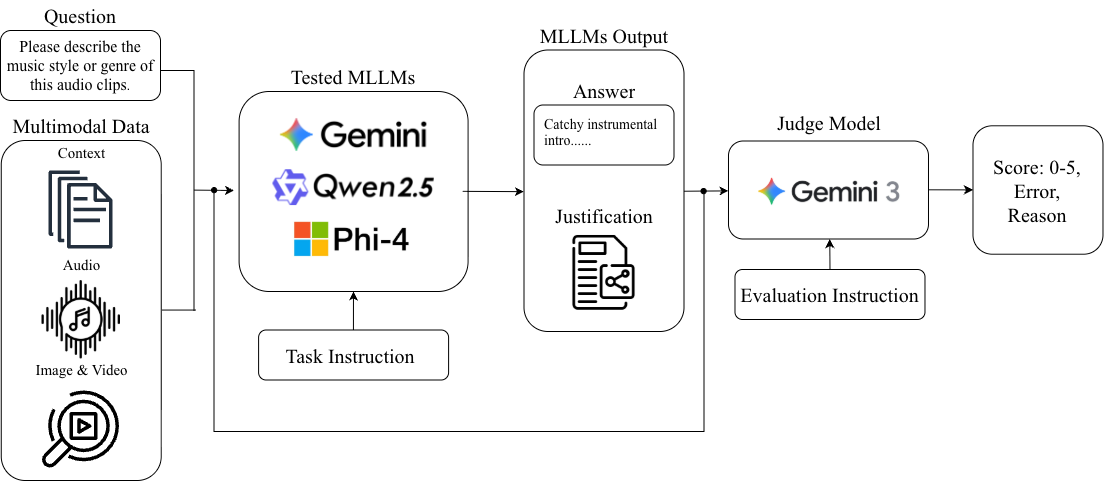}
    \caption{System architecture of the MLLMs evaluation framework. The framework takes multimodal inputs and task instructions to query multiple tested MLLMs. Each model produces an answer along with its justification. A judge model, following evaluation instructions, assesses the responses and outputs scores (0–5) and error analyses.}
    \label{fig:sys_infra}
\end{figure*}
\subsection{Overview of Evaluation Pipeline}
We design a scalable automated evaluation pipeline that assesses the capabilities of MLLMs across text, audio, image and video tasks. As illustration in Fig.~\ref{fig:sys_infra}, there are three main components: Multimodal data, Tested MLLMs, and Judge model. For each instance, the system retrieves a sample containing a query and its corresponding context. These inputs are fed into the tested MLLM, which generates a text-only response and a justification.

While our framework is architecturally designed to support a feedback loop--where errors are returned to the model for regeneration--this project focuses on validating the reliability of the Judge Model. Therefore, in our experiments, the pipeline works as a comprehensive forward-pass evaluation. That is, the Judge Model assesses the generated outputs against ground truth references and the multimodal evidence, and gives its own feedback.
\subsection{Tested MLLMs}
To ensure our benchmark covers a diverse range of capabilities, we evaluate three MLLMs that represent different model families and sizes.
\vspace{-0.5em}
\begin{itemize}[leftmargin=*]\itemsep -0.1em
    \item \textbf{Gemini-2.5}: An efficient multimodal model optimized for speed and reasoning.
    \item \textbf{Phi-4}: A powerful open-weight model developed by Microsoft, capable of handling complex tasks.
    \item \textbf{Qwen-2.5}: An omni-modal model designed to process and reason over various modality inputs.
\end{itemize}

\subsection{Judge Model}
The core of our methodology is the Judge model, powered by \textbf{Gemini-3-pro}. Unlike traditional text-only judges, this agent is explicitly prompted to perform multimodal reasoning. It receives the original task instruction, the multimodal context, the ground truth, and the tested model’s output. The Judge model has three primary objectives: \\
1. \textbf{Response Quality Analysis:} It determines if the final answer matches the ground truth or a physically observable reality.\\
2. \textbf{Reasoning Verification:} It analyzes the "Justification" provided by the tested model to check for logical consistency and hallucination. \\
3. \textbf{Diagnostic Feedback:} Instead of a simple binary pass/fail, the agent outputs a structured evaluation containing a numerical score (0-5), the specific error type, and a detailed reason for the score.
\vspace{-0.5em}
\subsection{Evaluation Metrics}
To quantify model performance, we utilize a rigorous Likert-scale scoring rubric. The Judge model assigns a score from 0 to 5 based on the following criteria defined in Table~\ref{tab:eval}. Beyond numerical scoring, the protocol includes a qualitative error analysis. As shown in our case studies, the Judge identifies specific failure modes, such as visual hallucinations or auditory omissions.

\begin{table}[h!]
\centering
\small
\resizebox{\linewidth}{!}{
    \begin{tabular}{c p{8cm}}
    \toprule
    \textbf{Score} & \textbf{Description} \\
    \midrule
    \rowcolor{green!70} \textbf{5} & Correct answer with sound justification. \\
    \rowcolor{green!40} \textbf{4} & Mostly correct answer and justification and minor errors. \\
    \rowcolor{yellow!70} \textbf{3}& Partially correct answer and justification, but errors exist. \\
    \rowcolor{yellow!40} \textbf{2}& Partially correct answer but unreasonable justification. \\
    \rowcolor{red!40}   \textbf{1} & Mostly wrong answer with unreasonable justification. \\
    \rowcolor{red!70}   \textbf{0} & Totally wrong answer or empty response. \\
    \bottomrule
    \end{tabular}
}
\caption{Evaluation metrics for the judge model and human judge to quantify tested MLLMs performance.}
\label{tab:eval}
\end{table}
\vspace{-1.8em}
\subsection{Experiment Results}
 Table~\ref{tab:exp} summarizes the average scores assigned by three human annotators and our Judge Model across four modalities and three MLLMs. Overall, we observe that tasks involving audio and video tend to achieve higher scores than those involving text-only or image inputs when averaging across all models, suggesting that current MLLMs may already possess relatively mature capabilities for temporal and acoustic understanding under our benchmark setting. In contrast, image-based and especially text-only tasks expose larger performance gaps between models: while the strongest system maintains scores close to 4, weaker models often fall near or below 2, indicating substantial room for improvement in fine-grained reading comprehension and visually grounded reasoning. \\
 From a model-wise perspective, Gemini-2.5 exhibits the most robust cross-modal performance. Its scores remain consistently high across all four modalities in most cases, with human and Judge ratings clustered roughly between 3.8 and 4.5, and without any catastrophic failure mode in a particular modality. This pattern suggests that Gemini’s internal representations transfer well between text, audio, image, and video inputs, and that its justifications are generally judged correct and well-grounded. In other words, under our evaluation protocol, Gemini behaves like a genuinely all-round multimodal model rather than a specialist tuned on a single dominant modality. \\
 In contrast, Phi-4 and Qwen-2.5 display more pronounced modality-specific weaknesses. Phi-4 delivers mid-range performance on audio and video, but its scores drop sharply on image tasks, where hallucinated visual details and misinterpreted chart elements frequently lead to scores near 2. Qwen-2.5 shows a complementary pattern: it performs competitively in audio and video, but its text-only scores are the lowest among all models, reflecting frequent failures in straightforward extraction and reasoning, including false refusals on questions with explicitly stated answers. Overall, these results indicate that cross-modal robustness is far from uniform across current MLLMs: some models generalize smoothly across modalities, while others remain highly sensitive to the dominant modality and the type of reasoning required.
\begin{table}[h!]
\centering
\scriptsize
\setlength{\tabcolsep}{2pt}
{
\renewcommand{\arraystretch}{0.5}
\newcolumntype{C}{>{\centering\arraybackslash}X}

\begin{tabularx}{\columnwidth}{@{} l l C C C C @{}}
\toprule
 & & \textbf{Human 1} & \textbf{Human 2} & \textbf{Human 3} & \textbf{Judge} \\
\midrule
\multirow{3}{*}{\textbf{Text}} 
& \textit{Gemini-2.5} & $\underline{3.98}$ & $\underline{\textbf{4.44}}$ & $\underline{4.12}$ & $\underline{4.28}$ \\
& \textit{Phi-4}            & $2.24$ & $2.12$ & $2.98$ & $2.45$ \\
& \textit{Qwen-2.5}    & $1.37$ & $1.28$ & $1.5$  & $1.15$ \\
\midrule

\multirow{3}{*}{\textbf{Audio}} 
& \textit{Gemini-2.5} & $\underline{3.58}$ & $3.62$ & $\underline{\textbf{4.16}}$ & $\underline{4.04}$ \\
& \textit{Phi-4}            & $2.66$ & $3.26$ & $3.30$ & $2.82$ \\
& \textit{Qwen-2.5}    & $3.26$ & $\underline{3.78}$ & $3.56$ & $3.38$ \\
\midrule

\multirow{3}{*}{\textbf{Video}} 
& \textit{Gemini-2.5} & $\underline{3.70}$ & $\underline{4.10}$ & $\underline{4.20}$ & $\underline{\textbf{4.54}}$ \\
& \textit{Phi-4}            & $2.78$ & $3.60$ & $3.50$ & $3.12$ \\
& \textit{Qwen-2.5}    & $2.82$ & $3.50$ & $3.02$ & $3.06$ \\
\midrule

\multirow{3}{*}{\textbf{Image}} 
& \textit{Gemini-2.5} & $\underline{3.46}$ & $\underline{3.82}$ & $\underline{\textbf{3.98}}$ & $\underline{3.94}$ \\
& \textit{Phi-4}            & $2.20$ & $2.66$ & $1.80$ & $2.24$ \\
& \textit{Qwen-2.5}    & $2.52$ & $3.28$ & $2.28$ & $2.32$ \\
\bottomrule
\end{tabularx}
}
\caption{Evaluation averaged score of Gemini-2.5, Phi-4, and Qwen-2.5 by three human annotators and the judge model, Gemini-3-pro, across multi-modal tasks.}
\label{tab:exp}
\end{table}
\vspace{-2.5em}
\section{Analysis}
\subsection{Score Distribution}
\begin{figure}[H]
    \centering
    \begin{subfigure}[b]{0.9\linewidth}
        \centering
        \includegraphics[width=\linewidth]{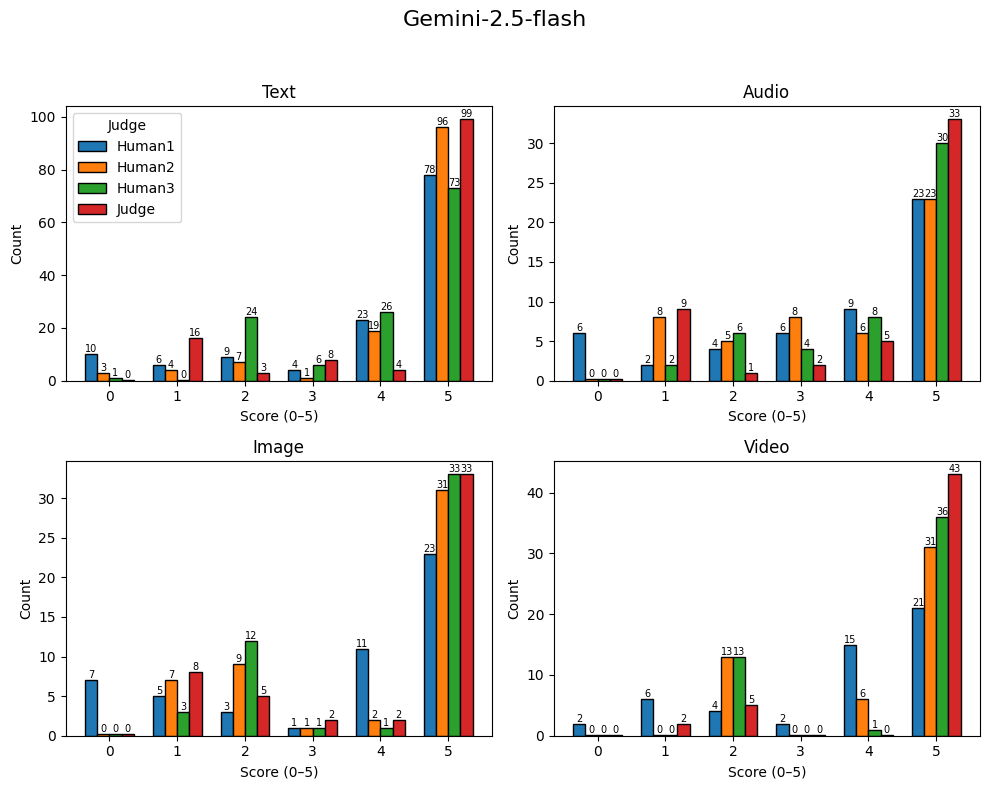}
        \label{fig:gemini}
    \end{subfigure}
    \begin{subfigure}[b]{0.9\linewidth}
        \centering
        \includegraphics[width=\linewidth]{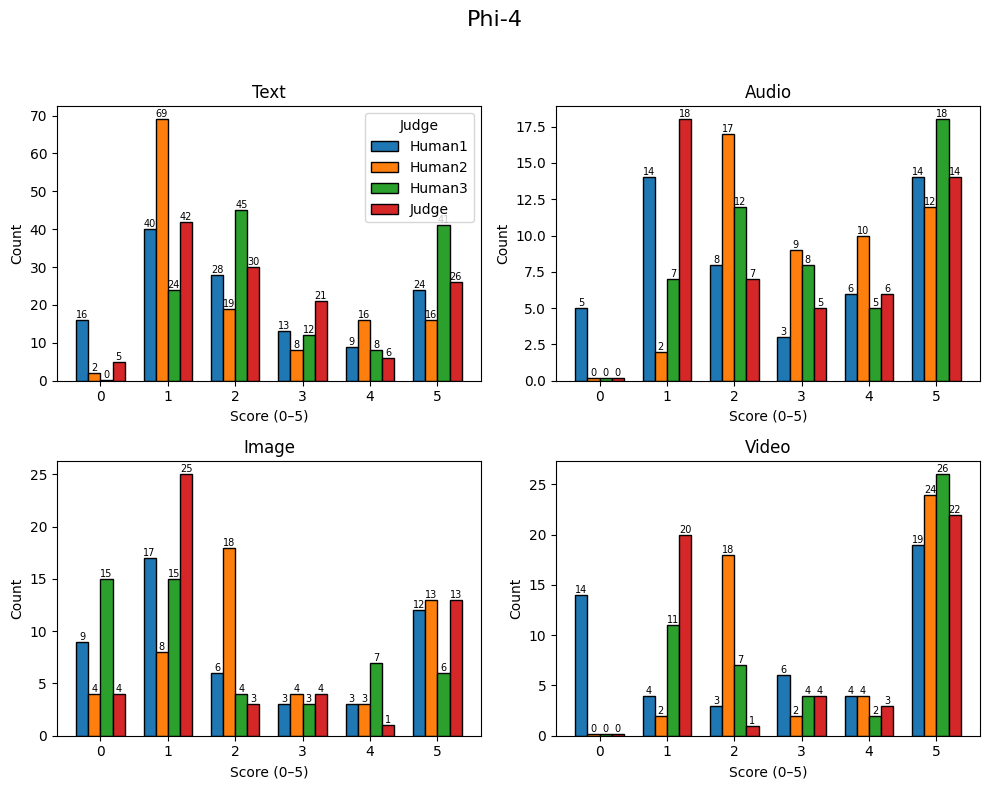}
        \label{fig:phi4}
    \end{subfigure}
    \begin{subfigure}[b]{0.9\linewidth}
        \centering
        \includegraphics[width=\linewidth]{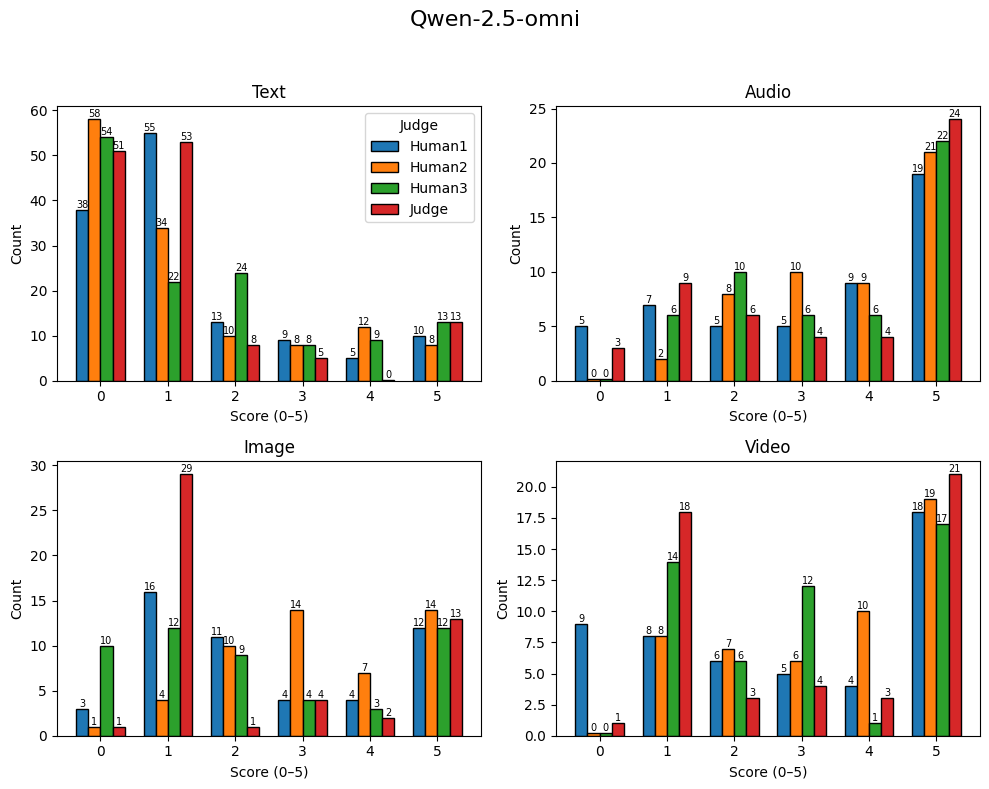}
        \label{fig:qwen}
    \end{subfigure}
    \caption{Human vs. Judge alignment across MLLMs.}
    \label{fig:human_judge_alignment}
\end{figure}

Figure~\ref{fig:human_judge_alignment} visualizes the score distributions of the three human annotators and the Judge Model for each model, complementing the averaged scores reported in Table~\ref{tab:exp}. Across Gemini-2.5, Phi-4, and Qwen-2.5, we observe that the Judge’s curve closely tracks the human trends in all four modalities: systems that receive higher mean scores from humans also receive higher scores from the Judge, and the relative ordering of models is preserved within each modality. At the same time, the Judge is slightly more conservative in its absolute calibration—its scores are typically lower than the three-annotator average by about 0.1–0.3 points. This systematic offset reflects the Judge’s stricter penalty on inconsistencies between the final answer and the provided justification, such as partially correct answers supported by hallucinated or logically incomplete reasoning. Despite this mild negative bias, the near-perfect agreement in rank ordering suggests that our automatic evaluator is reliable for comparative assessment of MLLMs, even if small corrections may be needed when interpreting its Likert scores in isolation.  

\subsection{Case Study}
\subsubsection{Text}

We illustrate a representative text QA example where the model must identify Carolina’s quarterback from a short American-football game recap. The paragraph explicitly states that “QB Jake Delhomme completed a 3-yard TD pass to WR Muhsin Muhammad,” and the question is ``who is carolina's quarterback?''. As shown in Table~\ref{tab:text_case}, Gemini-2.5 correctly extracts \emph{Jake Delhomme} and receives a perfect score from the Judge. In contrast, Qwen-2.5 incorrectly claims that the question cannot be answered, even though the answer is explicitly stated in the context, leading the Judge to label this case as a \emph{false refusal} with a low score. This example highlights that our Judge not only checks final accuracy but also penalizes missed evidence when the ground-truth answer is observable in the input.
\begin{table}[h!]
\centering
\scriptsize
\setlength{\tabcolsep}{3pt}
{
\renewcommand{\arraystretch}{0.9}
\begin{tabularx}{\columnwidth}{@{} l X c c @{}}
\toprule
\textbf{Model} & \textbf{Prediction} & \textbf{Score} & \textbf{Error Type} \\
\midrule
\textit{Gemini-2.5}
& Jake Delhomme 
& 5 
& None \\
\textit{Qwen-2.5}
& No, it's impossible to answer this question. 
& 1 
& False refusal \\
\bottomrule
\end{tabularx}
}
\caption{Text QA case study: identifying Carolina's quarterback from a game recap paragraph.}
\label{tab:text_case}
\end{table}
\vspace{-1.5em}
\subsubsection{Image}

\begin{figure}[h!]
    \centering
    \includegraphics[width=\linewidth]{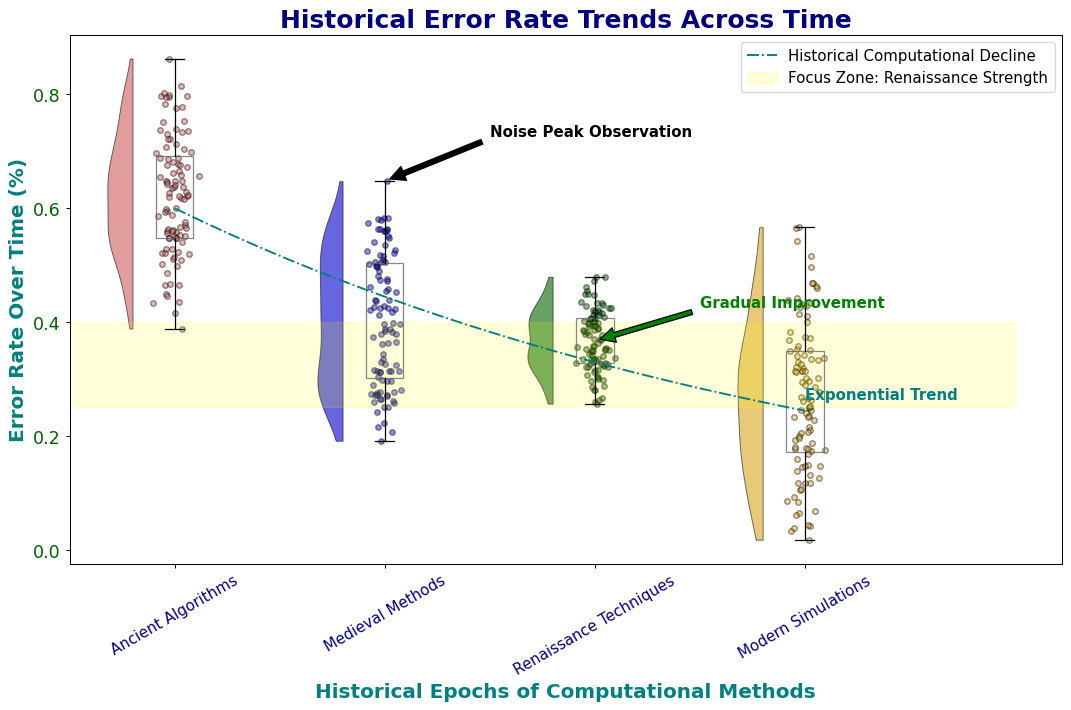}
    \caption{The input chart example for image case study.}
    \label{fig:img_case}
\end{figure}

For the image case study, we consider a chart-understanding question where the model must read the x-axis of a synthetic timeline plot shown in Fig.~\ref{fig:img_case} and answer: ``How many labeled epochs are present on the x-axis, and what are their names?''. The ground-truth chart contains four epochs: \emph{Ancient Algorithms}, \emph{Medieval Methods}, \emph{Renaissance Techniques}, and \emph{Modern Simulations}. As summarized in Table~\ref{tab:image_case}, Gemini-2.5 correctly recovers both the count and all four labels, and is therefore assigned the maximum score by the Judge. In contrast, Phi-4 also predicts that there are four epochs but hallucinates an entirely different set of names (e.g., \emph{Early Adopters}, \emph{Historical Adopters}), which do not appear anywhere in the chart. The Judge accordingly treats this as a hallucination case and assigns only partial credit. This example shows that our evaluator is sensitive not only to coarse numerical correctness but also to fine-grained semantic grounding in the visual evidence.

\begin{table}[h!]
\centering
\scriptsize
\setlength{\tabcolsep}{3pt}
{
\renewcommand{\arraystretch}{0.9}
\begin{tabularx}{\columnwidth}{@{} l X c c @{}}
\toprule
\textbf{Model} & \textbf{Prediction} & \textbf{Score} & \textbf{Error Type} \\
\midrule
\textit{Gemini-2.5}
& There are 4 labeled epochs: Ancient Algorithms, Medieval Methods, Renaissance Techniques, and Modern Simulations.
& 5 
& None \\
\textit{Phi-4} 
& The x-axis has 4 labeled epochs: Early Adopters, Historical Adopters, General Adoption, and Late Adopters.
& 3 
& Hallucination \\
\bottomrule
\end{tabularx}
}
\caption{Image case study: reading the labeled epochs on the x-axis of a chart.}
\label{tab:image_case}
\end{table}
\vspace{-1.5em}

\section{Future Work}
A promising direction for future work is to repurpose our Judge outputs as supervision signals. Each instance already includes a scalar score, an error type, and a natural-language explanation, which together form a rich signal that can be used as a reward model for reinforcement learning from LLM feedback or as a set of soft constraints in a Constitutional AI framework. Rather than stopping at offline evaluation, we could close the loop by fine-tuning MLLMs to maximize Judge scores and reduce specific failure modes such as hallucination and false refusal. Extending this idea beyond single-turn QA, future work may apply the same feedback mechanism to multi-turn agentic settings, where the Judge evaluates entire interaction traces instead of only final answers. Such a scaled-up pipeline would turn our Judge from static into active, enabling dynamically “learning from LLMs”.

\section{Conclusion}
We introduced a multimodal benchmark\footnote{\footnotesize \texttt{https://github.com/oscar-shih/multimodal\_judge\_benchmark}\\ \texttt{https://huggingface.co/multi-judge}} with a Judge for evaluating MLLMs across text, audio, image, and video tasks, combining Likert-scale scores with structured error types and natural-language explanations. Our experiments on Gemini-2.5, Phi-4, and Qwen-2.5, show that the Judge’s ratings closely track human annotators while being slightly more conservative, and reliably preserve the relative ranking of systems across modalities. Case studies further demonstrate that the Judge can distinguish between correct, hallucinated, and falsely refused answers in a fine-grained manner. 
\clearpage
\bibliography{anthology,custom}
\bibliographystyle{acl_natbib}




\end{document}